\def\year{2022}\relax
\documentclass[letterpaper]{article} % DO NOT CHANGE THIS
\usepackage{aaai22}  % DO NOT CHANGE THIS
\usepackage{times}  % DO NOT CHANGE THIS
\usepackage{helvet}  % DO NOT CHANGE THIS
\usepackage{courier}  % DO NOT CHANGE THIS
\usepackage[hyphens]{url}  % DO NOT CHANGE THIS
\usepackage{graphicx} % DO NOT CHANGE THIS
\usepackage{natbib}  % DO NOT CHANGE THIS AND DO NOT ADD ANY OPTIONS TO IT
\usepackage{caption} % DO NOT CHANGE THIS AND DO NOT ADD ANY OPTIONS TO IT
\usepackage{multirow}
\usepackage{amsfonts}
\usepackage{bm}
\usepackage{amsmath}
\newtheorem{definition}{Definition}
\DeclareCaptionStyle{ruled}{labelfont=normalfont,labelsep=colon,strut=off} % DO NOT CHANGE THIS

\usepackage{algorithm}
\usepackage{algorithmic}

\usepackage{newfloat}
\usepackage{listings}
\floatstyle{ruled}
\newfloat{listing}{tb}{lst}{}
\floatname{listing}{Listing}

\title{On the Effectiveness of Interpretable Feedforward Neural Network}
\author{
Miles Q. Li, Benjamin C. M. Fung\footnote{Corresponding author}, Adel Abusitta}
\affiliations {
    McGill University, Montreal, Canada\\
    miles.qi.li@mail.mcgill.ca, ben.fung@mcgill.ca, adel.abusitta@mcgill.ca
}

\begin{document}

\maketitle

\begin{abstract}
Deep learning models have achieved state-of-the-art performance in many classification tasks. However, most of them cannot provide an interpretation for their classification results. Machine learning models that are interpretable are usually linear or piecewise linear and yield inferior performance. Non-linear models achieve much better classification performance, but it is hard to interpret their classification results. This may have been changed by an interpretable feedforward neural network (IFFNN) proposed that achieves both high classification performance and interpretability for malware detection. If the IFFNN can perform well in a more flexible and general form for other classification tasks while providing meaningful interpretations, it may be of great interest to the applied machine learning community. In this paper, we propose a way to generalize the interpretable feedforward neural network to multi-class classification scenarios and any type of feedforward neural networks, and evaluate its classification performance and interpretability on intrinsic interpretable datasets. We conclude by finding that the generalized IFFNNs achieve comparable classification performance to their normal feedforward neural network counterparts and provide meaningful interpretations. Thus, this kind of neural network architecture has great practical use.
\end{abstract}

\section{Introduction}
Deep learning models are achieving state-of-the-art performance in an increasing number of tasks~\cite{jiang2019smart,foret2020sharpness,brown2020language,zhou2021informer}. They work as black-boxes, in which when a large number of training samples are fed to them, they learn patterns that correlate with different classes that are then used to classify unseen samples. However, most deep neural networks only implicitly learn and use the patterns, and do not explicitly explain the reasons for which a sample belongs to a class. This causes concerns about applying deep learning in some critical fields, such as healthcare and automatic pilot systems~\cite{choi2016retain,molnar2020interpretable,das2020opportunities,linardatos2021explainable}.

That being said, there are interpretable machine learning classification models, such as linear regression, softmax regression, and decision trees~\cite{laurent1976constructing}. These models can explain their classification results in a clear and simple way. However, as linear or piecewise linear models, their expressive abilities are very limited, i.e., they cannot model complex interactions between different features. Linear regression and softmax regression can be seen as neural networks with no hidden layers. They can tell to what extent each feature contributes to a classification result. The interpretability comes from that fact that the relation between a feature and the class of a sample is computed independently without any interactions. Even though this simplicity allows the model to explain its classification results, it yields inferior results compared to multi-layer neural networks. In this era, classification performance typically has higher priority than interpretability. Hence, these simple models are less useful than the complex and non-interpretable models~\cite{molnar2020interpretable}.

In an attempt to solve the dilemma of choosing either high classification performance or interpretability, some techniques have been proposed to interpret the classification results of complex machine learning models. For example, integrated gradients~\cite{sundararajan2017axiomatic} and permutation feature importance~\cite{datta2016algorithmic,koh2017understanding,adler2018auditing} can interpret many kinds of machine learning models. However, the model needs to be run many times to explain one prediction. The computational cost for an interpretation is too expensive. Some others, such as surrogate model methods~\cite{su2015interpretable,ribeiro2016nothing}, use another interpretable model (e.g., a decision tree) as a surrogate to approximate the target model and use the surrogate model's interpretation to explain the target model's prediction. However, the expressive ability of a surrogate model is usually not as good as a complex target model; thus, the former cannot very accurately approximate the latter, and the interpretation also cannot be accurate.

To avoid the aforementioned methods, some researchers have turned to deep neural networks that are self-explanatory. ~\cite{choi2016retain} propose RETAIN for classifying sequential data with an interpretation of how much each variable in the sequences contributes to the classification result.~\cite{LFCD21cose} propose an interpretable feedforward neural network (IFFNN) for malware detection. It classifies vectorial data and provides an interpretation of how much each feature in the vector contributes to the classification result. The proposed IFFNN architecture is promising for solving the dilemma between classification performance and interpretability. However, this exploration of IFFNN is very limited because it was only applied to binary classification, and the architecture contains fully connected layers and only accepts vectors as its input. In addition, the classification performance and interpretability were not comprehensively evaluated on general classification problems.

To explore whether the IFFNN proposed by ~\cite{LFCD21cose} can be extended to general classification scenarios and achieve excellent classification and meaningful interpretation, in this paper we generalize IFFNN to multi-class classification and any type of feedforward neural networks, and perform a comprehensive evaluation on the classification performance and interpretability of two interpretable datasets. The contributions of this paper are summarized as follows:
\begin{itemize}
\item We propose ways to generalize the IFFNN to multi-class classification and any type of feedforward neural network that takes any fixed shape tensor as its input.
\item We conduct comprehensive experiments to evaluate the classification performance and interpretability of the IFFNNs. We compare the classification accuracy of IFFNNs with their non-interpretable counterparts to show that they have similar classification performance.
\item We propose an interpretability benchmark dataset to evaluate the interpretability of classification models. It can generate an unlimited number of samples with the reasons why they belong to a specific class.
\end{itemize}

\section{Related Works}
\label{sec:rel}
Interpretability for machine learning models is acquired in different ways. For linear or piecewise linear models, such as linear regression, softmax regression, decision trees, and k-nearest neighbors, their simple classification mechanics make them intrinsically interpretable. Their expressive ability is quite limited so that they cannot achieve mundane classification performance when the features have complex interactions~\cite{molnar2020interpretable,das2020opportunities,linardatos2021explainable}.

Most complex machine learning models are not easily interpretable in themselves. Some post-hoc interpretation techniques have been proposed to interpret their classification results. Some interpretation methods do not require knowledge of the models. They just need the input and output pairs of the models to provide an interpretation. The permutation feature importance method~\cite{datta2016algorithmic,koh2017understanding,adler2018auditing} is one example of a model-agnostic method. The values of the features are permuted and then their impact on the classification results give a clue to how important they are. The computational cost is high since a model needs to be run multiple times. Surrogate model methods~\cite{su2015interpretable,ribeiro2016nothing} train an interpretable model, such as a decision tree, to approximate the target model to interpret, and use the interpretations given by the surrogate models to interpret the results given by the target model. As the expressive abilities of the surrogate models are usually lower than the target models, neither the approximation nor the interpretations are accurate. There are other interpretation techniques that work in a model-agnostic manner~\cite{lundberg2017unified,amoukou2021shapley}.

Other techniques are proposed to interpret certain types of machine learning models. The integrated gradients method~\cite{sundararajan2017axiomatic} is proposed to explain the classification results of neural networks (i.e., differentiable models) by cumulating the gradients along the path from a base sample to the target sample. Since this also requires running the target model multiple times, its efficiency is still limited. The fuzzy rule extraction method is proposed especially for interpretation of classification results for support vector machines~\cite{chaves2005fuzzy}. Other interpretation techniques are proposed for different types of neural networks, such as feedforward neural networks~\cite{LFCD21cose}, recurrent neural networks~\cite{choi2016retain,wisdom2016interpretable}, and convolutional neural networks~\cite{zeiler2010deconvolutional,zeiler2011adaptive,zeiler2014visualizing}.

\section{Problem Definition}
\label{sec:def}
Interpretability can be defined in different ways. To clarify the interpretability we discuss in this paper, we give the following formal definition of interpretation in a classification problem.

\begin{definition}[Interpretation]\label{defn:problem}
Let a sample be a $p$-th-order tensor $\bm{X}\in\mathbb{R}^{m_1\times m_2...\times m_p}$. The sample belongs to one of $c$ classes. An interpretable classification model should predict its class $y\in \{1,2,..,c\}$ and give an interpretation $\bm{I}\in\mathbb{R}^{c\times m_1\times m_2...\times m_p}$. $\bm{I}_{j,i_1,i_2,...,i_p}$ represents the importance of feature $\bm{X}_{i_1,i_2,...,i_p}$ for classifying it to class $j$.
\end{definition}

As can be seen from the definition of interpretation, it provides the importance value of a feature not only for the predicted class, but also for other classes. In practice, the interpretation does not have to be organized as a tensor $\bm{I}$. As long as an importance score of each element in $\bm{X}$ for each class can be computed, it is equivalent to having $\bm{I}$.

\section{Interpretable Feedforward Neural Network}
\label{sec:iffnn}
The interpretable feedforward neural network proposed by~\cite{LFCD21cose} contains a series of fully connected layers, which is similar to a normal feedforward neural network. The difference is that the output of the top layer $\bm{w}(\bm{x})$ is a vector that has the same dimension as the input feature vector and is used as a dynamically computed weight for the features. The last step is the same as logistic regression, which uses the dot product of the $\bm{w}(\bm{x})$ and $\bm{x}$, followed by sigmoid as the probability that a sample is positive.

The full computation is as follows. Let $\bm{x}\in \mathbb{R}^m$ be the feature vector of a sample. It is fed to $l$ fully connected hidden layers:

\begin{align}
\bm{v}_{l}(\bm{x}) &= FC^{l}(...FC^{1}(\bm{x})...)\\
where~FC^{i}(\bm{v}_{i-1}(\bm{x}))&=f(\bm{W}_1^i\bm{v}_{i-1}(\bm{x})+\bm{b}_1^i)
\end{align}
where $\bm{W}_1^i\in\mathbb{R}^{d_h^i\times d_h^{i-1}}$, $\bm{b}_1^i\in\mathbb{R}^{d_{h}^i}$, $f$ is the activation function (e.g., $Relu$, $tanh$), and $\bm{v}_{l}(\bm{x})\in\mathbb{R}^{d_{h}^l}$. Another normal fully connected layer where the output vector has the same dimension as $\bm{x}$ is applied:
\begin{equation}
\bm{w}(\bm{x}) = \bm{W_2}\bm{v}_{l}(\bm{x})+\bm{b_2} \label{bicls1}
\end{equation}
where $\bm{W_2}\in\mathbb{R}^{m\times d_{h}^l}$, $\bm{b_2}\in\mathbb{R}^{m}$, and $\bm{w}(\bm{x})\in\mathbb{R}^{m}$. $\bm{w}(\bm{x})$ serves as a weight vector for each feature in $\bm{x}$. The final confidence that the input sample belongs to the positive class (in malware detection, positive means malicious) is calculated as follows:
\begin{align}
y = IFFNN(\bm{x}) =& \sigma(\bm{w}(\bm{x})^T\bm{x}+b)  \label{bicls2}\\
where~\sigma(z) =&\frac{1}{1+e^{-z}}, b\in \mathbb{R}
\end{align}

This IFFNN has the modelling ability of a non-linear model since $\bm{w}(\bm{x})$ is computed through a multi-layer feedforward neural network. The interpretability of it is like logistic regression: the importance (i.e., contribution) of feature $x_i$ for the positive class is calculated as $w(\bm{x})_ix_i$ and the importance of feature $x_i$ for the negative class is $-w(\bm{x})_ix_i$.

\section{Generalization of Interpretable Feedforward Neural Networks}
The IFFNN can be generalized in different ways to be a more versatile neural network architecture for additional classification scenarios. We describe our methods of generalization in this section.

\subsection{Generalization to Multi-class Classification}
The original IFFNN is proposed for binary classification. It works as a logistic regression function with "dynamically" computed weights. Thus, a generalization of the original IFFNN to multi-class classification is to make it a software regression with "dynamically" computed weights.

Let $c$ be the number of classes and $\bm{W} \in \mathbb{R}^{c\times m}$ be a parametric matrix. Softmax regression can be expressed as follows:
\begin{align}
\bm{y} =& softmax(\bm{W}\bm{x}+\bm{b})\\
where~softmax(\bm{z}) =&\frac{1}{\sum_{j=1}^ce^{z_j}}(e^{z_1},...,e^{z_c}),\bm{b}\in\mathbb{R}^c
\end{align}
The output is a vector of dimension $c$, and each element is the probability that the sample belongs to a class. Therefore, $W_{i,j}x_j$ is the contribution of feature $x_j$ to class $i$.

For a multi-class classification scenario, rather than mapping the output of the last fully connected layer to a vector of dimension $m$, in the generalized IFFNN, the last fully connected layer requires a tensor to map the feature vector to a matrix that has the shape $c\times m$.

The complete computation of the generalized IFFNN for multi-class classification can be expressed as follows:
\begin{align}
\bm{v}_{l}(\bm{x}) &= FC^{l}(...FC^{1}(\bm{x})...)\\
\bm{W}(\bm{x}) &= \bm{T}\bm{v}_{l}(\bm{x})+\bm{B_2} \label{mulcls1}\\
\bm{y} &= softmax(\bm{W}(\bm{x})\bm{x}+\bm{b})  \label{mulcls2}
\end{align}
where $\bm{T} \in \mathbb{R}^{c \times m\times d_{h}^l}$, $\bm{B_2}\in \mathbb{R}^{c\times m}$, $\bm{W}(\bm{x})\in \mathbb{R}^{c\times m}$, and $\bm{b}\in\mathbb{R}^{c}$. The importance of feature $x_i$ to class $j$ is $W(\bm{x})_{j,i}x_i$.

In practice, it is equivalent to replace the tensor $T$ with a matrix $\bm{W_2}\in\mathbb{R}^{(cm)\times d_{h}^l}$. This matrix maps $\bm{v}_{l}(\bm{x})$ to a vector of dimension $cm$, which can be reshaped to a matrix with the expected shape $c \times m$. The complete equivalent computation of the generalized  IFFNN for multi-class classification can be expressed as follows:

\begin{align}
\bm{v}_{l}(\bm{x}) &= FC^{l}(...FC^{1}(\bm{x})...)\\
\bm{W}(\bm{x}) &= Reshape(\bm{W_2}\bm{v}_{l}(\bm{x}),(c \times m))+\bm{B_2}\\
\bm{y} &= softmax(\bm{W}(\bm{x})\bm{x}+\bm{b})  
\end{align}
where the $Reshape(\bm{z},target~shape)$ operation re-organizes the elements of $\bm{z}$ to the target shape.

\subsection{Generalization to Any Feedforward Neural Networks With Any Tensor of Fixed Shape As Input}

The original IFFNN can only be applied on vectors of fixed dimensions and only includes fully connected layers. These two constraints can be removed to build more expressive feedforward neural networks for wider applications. Rather than being a vector of a fixed dimension, the constraint for the input should be a tensor of a fixed shape. Vectors as first-order tensors are the most commonly seen feature form. Matrices as second-order tensors are also common input to feedforward neural networks. Greyscale images serve as a good example of this type. Furthermore, RBG images can be represented as third-order tensors. The feedforward neural networks that classify these high order tensors also usually contain other kinds of layers beyond fully connected layers, such as convolutional layers and pooling layers. We describe how to handle the generalized situations as follows.

Let $\bm{X}\in \mathbb{R}^{m_1\times m_2...\times m_p}$ be an order $p$ tensor representing the features of a sample. Let $m=m_1\times m_2... \times m_p$. For binary classification, we have:
\begin{align}
\bm{v}(\bm{X}) &= f(\bm{X})\\
\bm{w}(\bm{X}) &= \bm{W_2}\bm{v}(\bm{X})+\bm{b_2} \label{bc-begin}\\
\bm{x'}&=flatten(\bm{X})\\
\bm{y} &= \sigma(\bm{w}(\bm{X})^T\bm{x'} +b)   \label{bc-end}
\end{align}
where $f$ represents an arbitrary feedforward neural network with any kind of layers, $\bm{v}(\bm{X})\in \mathbb{R}^{d}$, $\bm{W_2}\in\mathbb{R}^{m\times d}$, $\bm{b_2},\bm{x'}\in\mathbb{R}^{m}$, the $flatten$ operation re-organizes the elements of a tensor to a 1d array to form a vector, and $b\in\mathbb{R}$. The importance of feature $X_{i_1,...,i_p}$ to the positive class is $w(\bm{X})_ix_i'$ where $i=i_1\times i_2\times...\times i_p$.

For multi-class classification, we have:
\begin{align}
\bm{v}(\bm{X}) &= f(\bm{X})\\
\bm{W}(\bm{X}) &= Reshape(\bm{W_2}\bm{v}(\bm{X}),(c \times m))+\bm{B_2} \label{mc-begin}\\
\bm{x'}&=flatten(\bm{X})\\
\bm{y} &= softmax(\bm{W}(\bm{X})\bm{x'} +\bm{b})  \label{mc-end}
\end{align}
where $\bm{v}(\bm{X})\in \mathbb{R}^{d}$, $\bm{W_2}\in\mathbb{R}^{(cm)\times d}$, $\bm{B_2}\in\mathbb{R}^{c\times m}$, $\bm{x'}\in\mathbb{R}^{m}$, and $\bm{b}\in\mathbb{R}^c$. The importance of feature $X_{i_1,...,i_p}$ to class $j$ is $W(\bm{X})_{j,i}x_i'$ where $i=i_1\times i_2\times...\times i_p$.

It should be noted that assuming $\bm{v}(\bm{X})$, the output of $f(\bm{X})$ as a vector of a fixed dimension does not cause the loss of generality. When $f(\bm{X})$ is a higher order tensor rather than a vector, its shape is still fixed, so it can always be converted to a vector by applying a $flatten$ operation.

\subsection{Discussion}
In some cases, in the input tensor, multiple elements correspond to the same object. When the importance of each object is expected, the importance of these elements should be added up. For instance, an RGB image can be represented as a third-order tensor $\bm{X}\in\mathbb{R}^{3\times h \times w}$. $X_{0,i,j}$, $X_{1,i,j}$, and $X_{2,i,j}$ are the red, green, and blue values of the same pixel. Their importance of pixel $(i,j)$ is the summation of the importance values of $X_{0,i,j}$, $X_{1,i,j}$, and $X_{2,i,j}$.

\section{Experiments}
\label{sec:eval}
In this section, we evaluate various versions of IFFNNs on different datasets. The objectives are to answer the following questions:
\begin{itemize}
\item Is classification performance harmed when the feedforward neural networks are organized in our interpretable way compared to normal feedforward neural networks?
\item Do the interpretations given by the IFFNNs make sense?
\item Do the generalized versions of IFFNNs work well in terms of classification performance and interpretability?
\end{itemize}

\subsection{Datasets}

We evaluate the models on two datasets: MNIST and INBEN. They complement each other in the evaluation procedure. MNIST is an image classification dataset that allows us to evaluate IFFNNs with convolutional layers and to qualitatively evaluate the interpretability of IFFNNs. However, it cannot be used to quantitatively evaluate their interpretability, since there is no exact answer on how important each pixel is for the classification results. With our created dataset INBEN, the gold standard interpretations of the samples are known, and thus allows us to achieve this purpose.

\begin{table}[h]
\caption{Statistics of the datasets used for evaluation.}
\label{tab:datasets}
\begin{center}
\begin{tabular}{|c|c|c|c|c|}
\hline
Dataset&Training&Valid&Test&$X$ Shape\\\hline
MNIST 10 cls&50,000&10,000&10,000&(28,28)\\
MNIST 2 cls&10,554&2,111&2,115&(28,28)\\
INBEN 10 cls&100,000&10,000&10,000&(1000,)\\
INBEN 2 cls&20,000&2,000&2,000&(1000,)\\
\hline
\end{tabular}
\end{center}
\end{table}

\subsubsection{MNIST}
MNIST is a handwritten digit dataset. It is a common benchmark for image classification models. This dataset works well for our purposes because of its easily interpretable character. The IFFNNs applied on this dataset can point out which pixels are important to classify a sample to a certain digit. It is easy for humans to determine whether these pixels are good indicators for the predictions.

We create two scenarios with MNIST. \textbf{Scenario 1} uses samples on all 10 classes. This can be used to evaluate the generalized IFFNN to multi-class classification. \textbf{Scenario 2} uses samples of only two classes (digits of "0" and "1") which can evaluate both the binary classification versions and multi-class classification versions of IFFNNs.

\subsubsection{INBEN}

By visualizing the importance of each pixel of an image in MNIST, we can only qualitatively evaluate the interpretability of the IFFNNs. To quantitatively evaluate the interpretability, we propose an INterpretablility BENchmark (INBEN) dataset. It can be described as follows:

\begin{enumerate}
\item Each sample belongs to 1 of $c$ classes.
\item Each sample is a vector of dimension $m$. Each entry corresponds to a fixed feature, and the value of it could be 0 or 1. For example, if $m=5$, a sample could be (1 0 1 1 0).
\item For each class, there is a set of randomly generated patterns, where if a sample contains one of these patterns, it belongs to that class. For example, (1,3) is a pattern for class 2. It means that a sample $x$ belongs to class 2 if $x_1=1$ and $x_3=1$. (1 0 1 1 0) is an example that contains this pattern.
\item There is a class priority sequence (e.g., [3,2,4,1,0]). If a sample contains patterns of multiple classes, it belongs to the class with the highest priority among them. For example, if a sample contains the patterns of both class 2 and class 0, it belongs to class 2.
\item There is a default class. If a sample contains no patterns, it belongs to the default class.
\end{enumerate}

We also create two scenarios with INBEN datasets. \textbf{Scenario 1} contains 10 classes of samples, and \textbf{Scenario 2} contains 2 classes of samples. The latter can be used to evaluate the IFFNNs for binary classification as well.

The statistics of the datasets are given in Table~\ref{tab:datasets}. 

\begin{table*}[h]
\caption{Classification performance evaluation on MNIST and INBEN.}
\label{tab:cls_result}
\begin{center}
\begin{tabular}{|c|cc|cc|cc|cc|}
\hline
\multirow{2}{*}{Model}&\multicolumn{2}{|c|}{10-class MNIST}&\multicolumn{2}{|c|}{2-class MNIST}&\multicolumn{2}{|c|}{10-class INBEN}&\multicolumn{2}{|c|}{2-class INBEN}\\
&Params&Acc&Params&Acc&Params&Acc&Params&Acc\\\hline
FC-MC1&898.5K&98.46&894.5K&99.93&1.0M&97.80&1.0M&98.23\\
FC-MC2&4.8M&98.54&1.7M&99.94&6.0M&98.83&2.0M&98.45\\
FC-MC3&4.8M&98.49&1.7M&99.92&6.0M&98.69&2.0M&98.37\\
FC-IFFNN-MC&4.8M&98.06&1.7M&99.91&6.0M&98.19&2.0M&99.06\\\hline
HW-MC1&2.4M&98.13&2.4M&99.93&2.5M&97.99&2.5M&98.57\\
HW-MC2&6.3M&98.10&3.2M&99.92&7.5M&97.81&3.5M&98.69\\
HW-MC3&6.3M&97.67&3.2M&99.93&7.5M&97.41&3.5M&98.68\\
HW-IFFNN-MC&6.3M&97.96&3.2M&99.90&7.5M&97.58&3.5M&99.28\\\hline
ResNET-MC1&226.2K&99.50&201.1K&99.92&NA&NA&NA&NA\\
ResNET-MC2&24.7M&99.41&5.1M&99.99&NA&NA&NA&NA\\
ResNET-MC3&24.7M&99.39&5.1M&99.93&NA&NA&NA&NA\\
ResNET-IFFNN-MC&24.8M&98.92&5.1M&99.95&NA&NA&NA&NA\\\hline
CNN-MC1&1.2M&98.88&1.2M&99.89&NA&NA&NA&NA\\
CNN-MC2&72.3M&98.95&14.5M&99.92&NA&NA&NA&NA\\
CNN-MC3&72.3M&98.99&14.5M&99.93&NA&NA&NA&NA\\
CNN-IFFNN-MC&72.3M&98.69&14.5M&99.96&NA&NA&NA&NA\\\hline
SR&7.8K&92.82&1.6K&99.95&10.0K&87.53&2.0K&97.67\\\hline
DT &NA&88.19&NA&99.66&NA&76.75&NA&98.93\\\hline
FC-BC1&NA&NA&894.0K&99.95&NA&NA&1.0M&98.04\\
FC-BC2&NA&NA&1.3M&99.92&NA&NA&1.5M&98.47\\
FC-BC3&NA&NA&1.3M&99.91&NA&NA&1.5M&98.58\\
FC-IFFNN-BC&NA&NA&1.3M&99.94&NA&NA&1.5M&98.67\\\hline
HW-BC1&NA&NA&2.4M&99.92&NA&NA&2.5M&98.71\\
HW-BC2&NA&NA&2.8M&99.91&NA&NA&3.0M&98.55\\
HW-BC3&NA&NA&2.8M&99.92&NA&NA&3.0M&98.57\\
HW-IFFNN-BC&NA&NA&2.8M&99.94&NA&NA&3.0M&99.34\\\hline
ResNET-BC1&NA&NA&197.9K&99.98&NA&NA&NA&NA\\
ResNET-BC2&NA&NA&2.7M&99.95&NA&NA&NA&NA\\
ResNET-BC3&NA&NA&2.7M&99.96&NA&NA&NA&NA\\
ResNET-IFFNN-BC&NA&NA&2.7M&99.91&NA&NA&NA&NA\\\hline
CNN-BC1&NA&NA&1.2M&99.93&NA&NA&NA&NA\\
CNN-BC2&NA&NA&7.2M&99.93&NA&NA&NA&NA\\
CNN-BC3&NA&NA&7.2M&99.91&NA&NA&NA&NA\\
CNN-IFFNN-BC&NA&NA&7.2M&99.94&NA&NA&NA&NA\\\hline
LR&NA&NA&0.8K&99.95&NA&NA&1.0K&97.66\\\hline

\hline
\end{tabular}
\end{center}
\end{table*}

\subsection{Models}
We include four kinds of feedforward neural networks in our experiments to illustrate the classification performance and interpretability of the IFFNN. They are fully connected feedforward neural networks (FC), convolutional neural networks (CNN)~\cite{lecun1998gradient},  fully connected feedforward neural networks with highways (HW)~\cite{srivastava2015highway}, and residual neural networks (ResNET)~\cite{he2016deep}. For each of the four kinds of neural networks, we have eight different variants. We use FC as the example to describe the variants:
\begin{itemize}
\item  \textbf{FC-BC1} A feedforward neural network with fully connected layers for binary classification. The top fully connected layer maps the feature vector to a real number followed by a sigmoid layer. This is only applicable to \textbf{Scenario 2}.
\item  \textbf{FC-MC1} A feedforward neural network with fully connected layers for multi-class classification. The top fully connected layer maps the feature vector to a vector of dimension $c$ followed by a softmax layer. 
\item  \textbf{FC-IFFNN-BC} The interpretable version of FC-BC1 achieved by replacing the top layer with Eq.\ref{bc-begin}$\sim$\ref{bc-end}. This is only applicable to \textbf{Scenario 2}.
\item  \textbf{FC-IFFNN-MC} The interpretable version of FC-MC1 achieved by replacing the top layer with Eq.\ref{mc-begin}$\sim$ \ref{mc-end}.
\item  \textbf{FC-BC2} Similar to FC-BC1, with the total number of trainable parameters about the same as FC-IFFNN-BC by increasing the dimensions of the layers but not increasing the number of layers. This is only applicable to \textbf{Scenario 2}.
\item  \textbf{FC-MC2} Similar to FC-MC1, with the total number of trainable parameters about the same as FC-IFFNN-MC by increasing the dimensions of the layers but not increasing the number of layers.
\item  \textbf{FC-BC3} Similar to FC-BC1, with the total number of trainable parameters about the same as FC-IFFNN-BC by increasing the number of layers, and adjusting the dimension of each layer. This is only applicable to \textbf{Scenario 2}.
\item  \textbf{FC-MC3} Similar to FC-MC1, with the total number of trainable parameters about the same as FC-IFFNN-MC by increasing the number of layers, and adjusting the dimension of each layer.
\end{itemize}

For the other three kinds of neural networks, there are the same eight variants. When we apply FC and HW networks on the MNIST dataset, we flatten the input to a vector. We don't apply CNN and ResNET on INBEN because those two networks are mainly for input of matrices or third-order tensors.

We also compare with other interpretable models, including logistic regression (LR), softmax regression (SR), and decision trees (DT). We use grid search to tune the hyper-parameters of decision trees, including its split criterion and maximum depth. The candidate values are given in Table~\ref{tab:hyperdt}.

\begin{table}[h]
\caption{Candidate values for hyper-parameters of decision tree.}
\label{tab:hyperdt}
\begin{center}
\begin{tabular}{|c|c|}
\hline
Hyperparameter&Candidate Values\\\hline
Split Criterion&gini,entropy\\
Maximum Depth&10,25,50,100,200,300,400,500,1000\\
\hline
\end{tabular}
\end{center}
\end{table}

\subsection{Evaluation Metrics}
We describe the evaluation metrics for classification performance and interpretability in this section. 

For the classification performance, following the tradition, we use the metric of \textbf{accuracy}, which is the number of correctly classified samples over the total number of samples.

We cannot use MNIST to quantitatively evaluate the interpretability of the models, but we can use INBEN. With INBEN, we know the reason why a sample belongs to a class. It is the pattern(s) that decides its class. The ideal interpretations should give the features included in the patterns the greatest importance values. Therefore, we use the average of \textbf{accuracy@N} as our evaluation metric for interpretability. We formally define it as follows:

\begin{definition}[Accuracy@N]\label{defn:accatn}
Let $S_1$ be the set of features in the pattern(s) that determines a sample $\bm{x}$ belong to class $c$. Let $N=|S_1|$. Let $S_2$ be the set of top $N$ important features for classifying $\bm{x}$ to class $c$ by an interpretable classification system. Let $S_3=S_1\cap S_2$ and $n=|S_3|$. Then, $Accuracy@N=n/N$.
\end{definition}

As can be seen, N is variant to different samples. Below is an example.

A sample $\bm{x}$ belongs to class 2 because it contains the two patterns of class 2: (113,251) and (35,72,99,217,251). We thus have $S_1=\{35,72,99,113,217,251\}$ and $N=6$. The top six most important features for classifying it to class 2 determined by IFFNN are: 113,251,7,35,12,308,221. So, $S_2=\{7,12,35,113,221,251,308\}$. $S_3=\{35,113,251\}$ and thus $n=3$. $Accuracy@N=\frac{3}{6}=0.5$.

We use the average of accuracy@N over all correctly classified test samples as the evaluation metric for interpretability. We do not include wrongly classified samples because these do not mean anything.

\begin{figure*}
\centering
\includegraphics[width=14cm]{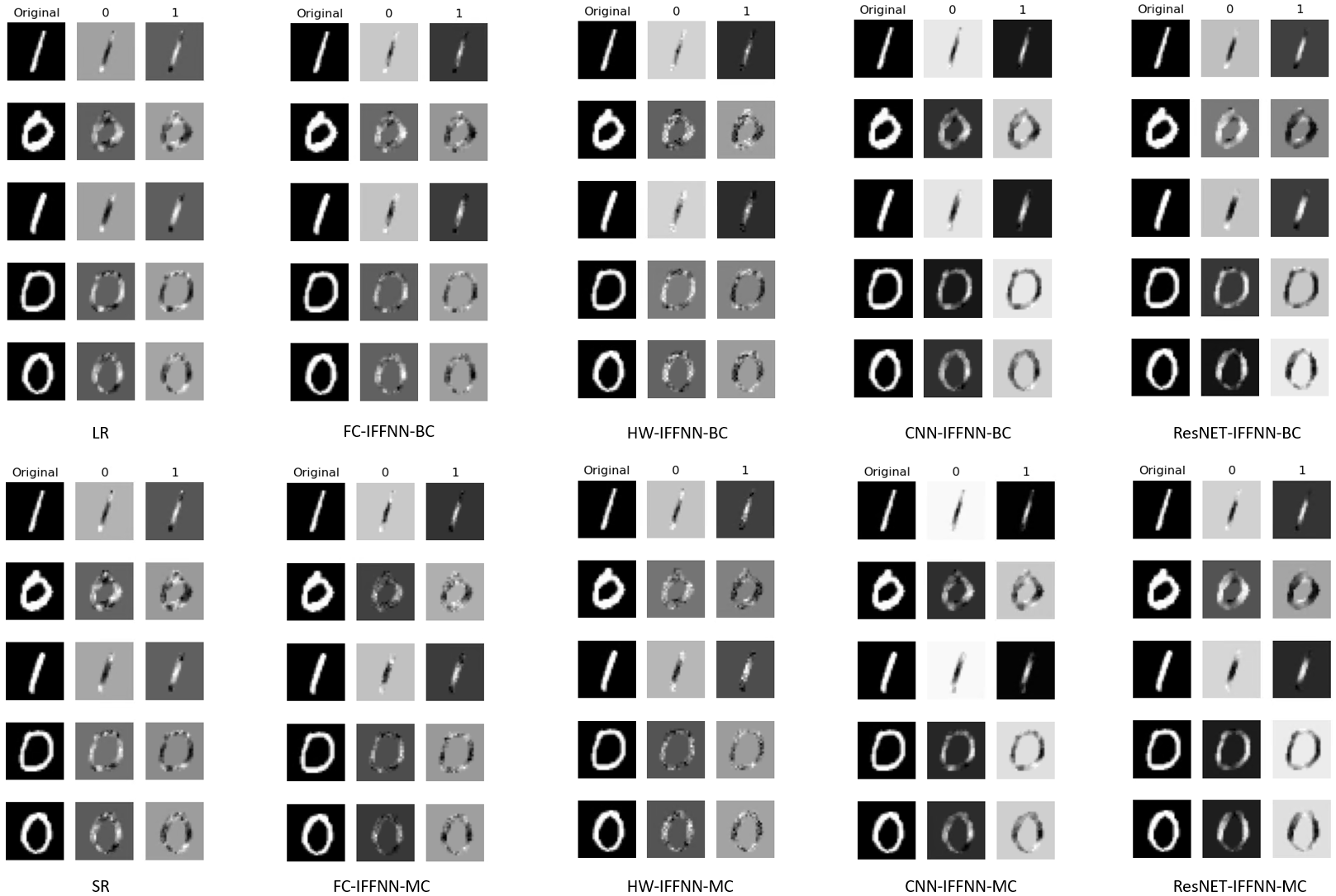}
 \caption{Examples of images and the interpretations for the classifications on MNIST with only 0 and 1.}
 \label{fig:inter2cls}
\end{figure*}

\subsection{Experiment Setting}
We train and evaluate the models on a server with two Xeon E5-2697 CPUs, 384 GB of memory, and four Nvidia Titan XP graphics cards. Only one graphics card is used for each run. The operating system is Windows Server 2016. We use Python 3.7.9 and PyTorch 1.6.0~\cite{paszke2017automatic} to implement the models. We use the implementation of DT in scikit-learn 0.23.2~\cite{scikit-learn}.

We use Adam~\cite{kingma2014adam} with the initial learning rate $1e-3$ to train all the neural networks including LR and SR. The batch size is 256 and maximum epoch is 200. The accuracy on the test set at the epoch in which the accuracy on the validation set is the best is reported.  

We repeat each group of experiments five times and report the average. To guarantee that our experiment results are 100\% reproducible, we use random seeds from 0 to 4 for model initializations.

\subsection{Classification Results}

The classification performance of all models is shown in Table~\ref{tab:cls_result}. The IFFNN version of different types of feedforward neural networks achieves slightly higher or lower accuracy compared with the non-interpretable ones in most cases (i.e., the difference is at most 1\%). Between the same kind of neural networks with different amounts of trainable parameters, the difference in accuracy is minor as well. On datasets with 10 classes of samples, we can see a significant gap ($>5\%$) between SR, DT, and the neural networks. This means that forming feedforward neural networks in the proposed interpretable way does not harm the classification performance and is as effective as a normal multi-layer neural network. The generalized versions of IFFNNs on different feedforward neural networks for multi-class classification also perform well in terms of accuracy. For the reason of limited space, we could not put too detailed information on the results in the manuscript.

\subsection{Interpretability Results}

\begin{table}[h]
\caption{Evaluation of interpretability with Accuracy@N on INBEN.}
\label{tab:inter}
\begin{center}
\begin{tabular}{|c|c|c|}
\hline
Model&10-class INBEN&2-class INBEN\\\hline
SR         &86.7\%&83.8\%\\
FC-IFFNN-MC&98.4\%&91.3\%\\
HW-IFFNN-MC&98.3\%&95.3\%\\\hline
LR         &NA&83.8\%\\
FC-IFFNN-BC&NA&90.5\%\\
HW-IFFNN-BC&NA&95.3\%\\
\hline
\end{tabular}
\end{center}
\end{table}

\subsubsection{Quantitative Analysis}

The Accuracy@N of LR, SR, and the IFFNNs on INBEN are reported in Table~\ref{tab:inter}. As shown, the Accuracy@N of IFFNNs is always larger than 90\%, which means when a sample is correctly classified, the IFFNNs can correctly point out the features in the patterns that determine its class. This indicates that the interpretations provided by them are meaningful. 

We can also see that the interpretations given by IFFNNs are even more accurate than those given by LR and SR. The reason is that the INBEN dataset we created is non-linear, thus these linear models cannot capture the patterns that determine the class of a sample. To be more specific, LR and SR can only model the relation between a feature and a class independently, however, the patterns require the models to be able to model the co-occurrences of different features. Multi-layer neural networks model interactions of different features through the computations in the hidden layers. This also reflects the fact that as multi-layer networks, the IFFNNs have the pattern recognition ability of non-linear models.

\subsubsection{Qualitative Analysis}
We qualitatively evaluate the models on MNIST. We show the importance of a pixel to a class in a greyscale image that has the same shape as the original image, and the greyscale of a pixel is the importance of the pixel in the same position. The greyscale of the background in the original images is always 0, so their importance is also 0. Therefore, the pixels that are lighter than the background provide a positive contribution to the class and the darker pixels provide a negative contribution. We use the scenario with only "0" and "1" as for the evaluation because there are areas of the images that only contain white pixels for only one of them and these pixels are good indicators of the digits.

Figure~\ref{fig:inter2cls} show some images from the test set and the importance images of them for all classes. It is clearer to qualitatively evaluate the interpretability from Figure~\ref{fig:inter2cls}. We can see that for the images of "0", the important pixels for the right class (i.e., "0") determined by all IFFNNs focus on the pixels of the left and right parts of the circle. This makes sense because "1" usually is close to a vertical bar, and white pixels rarely appear in those areas of the images of "1". Therefore, it makes sense that white pixels appearing in these areas contribute more to the class of "0". The important pixels for images of "1" are more concentrated in the center part of the stroke. This is also valid because there are rarely white pixels in the center areas of images of "0".

\section{Conclusion}
\label{sec:conc}
In this paper, we propose ways to generalize the IFFNN proposed in~\cite{LFCD21cose} to multi-class classification and any type of feedforward neural networks. We also conduct comprehensive experiments to evaluate the classification performance and interpretability of the IFFNNs. We reached the conclusion that the IFFNNs achieve similar classification accuracy as their non-interpretable feedforward neural network counterparts and provide meaningful interpretations. Therefore, the generalized IFFNN architecture is an excellent choice for real-world applications when interpretations for classification results are expected for various reasons.

\section*{Acknowledgment}

This research was funded by NSERC Discovery Grants (RGPIN-2018-03872), Canada Research Chairs Program (950-230623), and the Canadian National Defence Innovation for Defence Excellence and Security (IDEaS W7714-217794/001/SV1). The IDEaS program assists in solving some of Canada’s toughest defence and security challenges. The Titan Xp used for this research was donated by the NVIDIA Corporation.

\bibliography{aaai22}

\begin{thebibliography}{29}
\providecommand{\natexlab}[1]{#1}

\bibitem[{Adler et~al.(2018)Adler, Falk, Friedler, Nix, Rybeck, Scheidegger,
  Smith, and Venkatasubramanian}]{adler2018auditing}
Adler, P.; Falk, C.; Friedler, S.~A.; Nix, T.; Rybeck, G.; Scheidegger, C.;
  Smith, B.; and Venkatasubramanian, S. 2018.
\newblock Auditing black-box models for indirect influence.
\newblock \emph{Knowledge and Information Systems}, 54(1): 95--122.

\bibitem[{Amoukou, Brunel, and Sala{\"u}n(2021)}]{amoukou2021shapley}
Amoukou, S.~I.; Brunel, N.~J.; and Sala{\"u}n, T. 2021.
\newblock The Shapley Value of coalition of variables provides better
  explanations.
\newblock \emph{arXiv preprint arXiv:2103.13342}.

\bibitem[{Brown et~al.(2020)Brown, Mann, Ryder, Subbiah, Kaplan, Dhariwal,
  Neelakantan, Shyam, Sastry, Askell et~al.}]{brown2020language}
Brown, T.~B.; Mann, B.; Ryder, N.; Subbiah, M.; Kaplan, J.; Dhariwal, P.;
  Neelakantan, A.; Shyam, P.; Sastry, G.; Askell, A.; et~al. 2020.
\newblock Language models are few-shot learners.
\newblock \emph{arXiv preprint arXiv:2005.14165}.

\bibitem[{Chaves, Vellasco, and Tanscheit(2005)}]{chaves2005fuzzy}
Chaves, A.~C.; Vellasco, M.~M.; and Tanscheit, R. 2005.
\newblock Fuzzy rule extraction from support vector machines.
\newblock In \emph{Fifth International Conference on Hybrid Intelligent Systems
  (HIS'05)}, 6--pp. IEEE.

\bibitem[{Choi et~al.(2016)Choi, Bahadori, Kulas, Schuetz, Stewart, and
  Sun}]{choi2016retain}
Choi, E.; Bahadori, M.~T.; Kulas, J.~A.; Schuetz, A.; Stewart, W.~F.; and Sun,
  J. 2016.
\newblock Retain: An interpretable predictive model for healthcare using
  reverse time attention mechanism.
\newblock \emph{arXiv preprint arXiv:1608.05745}.

\bibitem[{Das and Rad(2020)}]{das2020opportunities}
Das, A.; and Rad, P. 2020.
\newblock Opportunities and challenges in explainable artificial intelligence
  (xai): A survey.
\newblock \emph{arXiv preprint arXiv:2006.11371}.

\bibitem[{Datta, Sen, and Zick(2016)}]{datta2016algorithmic}
Datta, A.; Sen, S.; and Zick, Y. 2016.
\newblock Algorithmic transparency via quantitative input influence: Theory and
  experiments with learning systems.
\newblock In \emph{2016 IEEE symposium on security and privacy (SP)}, 598--617.
  IEEE.

\bibitem[{Foret et~al.(2020)Foret, Kleiner, Mobahi, and
  Neyshabur}]{foret2020sharpness}
Foret, P.; Kleiner, A.; Mobahi, H.; and Neyshabur, B. 2020.
\newblock Sharpness-aware minimization for efficiently improving
  generalization.
\newblock \emph{arXiv preprint arXiv:2010.01412}.

\bibitem[{He et~al.(2016)He, Zhang, Ren, and Sun}]{he2016deep}
He, K.; Zhang, X.; Ren, S.; and Sun, J. 2016.
\newblock Deep residual learning for image recognition.
\newblock In \emph{Proceedings of the IEEE conference on computer vision and
  pattern recognition}, 770--778.

\bibitem[{Jiang et~al.(2019)Jiang, He, Chen, Liu, Gao, and
  Zhao}]{jiang2019smart}
Jiang, H.; He, P.; Chen, W.; Liu, X.; Gao, J.; and Zhao, T. 2019.
\newblock Smart: Robust and efficient fine-tuning for pre-trained natural
  language models through principled regularized optimization.
\newblock \emph{arXiv preprint arXiv:1911.03437}.

\bibitem[{Kingma and Ba(2014)}]{kingma2014adam}
Kingma, D.~P.; and Ba, J. 2014.
\newblock Adam: A method for stochastic optimization.
\newblock \emph{arXiv preprint arXiv:1412.6980}.

\bibitem[{Koh and Liang(2017)}]{koh2017understanding}
Koh, P.~W.; and Liang, P. 2017.
\newblock Understanding black-box predictions via influence functions.
\newblock In \emph{International Conference on Machine Learning}, 1885--1894.
  PMLR.

\bibitem[{Laurent and Rivest(1976)}]{laurent1976constructing}
Laurent, H.; and Rivest, R.~L. 1976.
\newblock Constructing optimal binary decision trees is NP-complete.
\newblock \emph{Information processing letters}, 5(1): 15--17.

\bibitem[{LeCun et~al.(1998)LeCun, Bottou, Bengio, and
  Haffner}]{lecun1998gradient}
LeCun, Y.; Bottou, L.; Bengio, Y.; and Haffner, P. 1998.
\newblock Gradient-based learning applied to document recognition.
\newblock \emph{Proceedings of the IEEE}, 86(11): 2278--2324.

\bibitem[{Li et~al.(2021)Li, Fung, Charland, and Ding}]{LFCD21cose}
Li, M.~Q.; Fung, B. C.~M.; Charland, P.; and Ding, S. H.~H. 2021.
\newblock {I-MAD}: Interpretable Malware Detector Using {Galaxy Transformers}.
\newblock \emph{Computers \& Security (COSE)}, 108(102371): 1--15.

\bibitem[{Linardatos, Papastefanopoulos, and
  Kotsiantis(2021)}]{linardatos2021explainable}
Linardatos, P.; Papastefanopoulos, V.; and Kotsiantis, S. 2021.
\newblock Explainable ai: A review of machine learning interpretability
  methods.
\newblock \emph{Entropy}, 23(1): 18.

\bibitem[{Lundberg and Lee(2017)}]{lundberg2017unified}
Lundberg, S.~M.; and Lee, S.-I. 2017.
\newblock A unified approach to interpreting model predictions.
\newblock In \emph{Proceedings of the 31st international conference on neural
  information processing systems}, 4768--4777.

\bibitem[{Molnar(2020)}]{molnar2020interpretable}
Molnar, C. 2020.
\newblock \emph{Interpretable machine learning}.
\newblock Lulu.com.

\bibitem[{Paszke et~al.(2017)Paszke, Gross, Chintala, Chanan, Yang, DeVito,
  Lin, Desmaison, Antiga, and Lerer}]{paszke2017automatic}
Paszke, A.; Gross, S.; Chintala, S.; Chanan, G.; Yang, E.; DeVito, Z.; Lin, Z.;
  Desmaison, A.; Antiga, L.; and Lerer, A. 2017.
\newblock Automatic differentiation in PyTorch.
\newblock \emph{Neural Information Processing Systems NIPS 2017 Autodiff
  Workshop}.

\bibitem[{Pedregosa et~al.(2011)Pedregosa, Varoquaux, Gramfort, Michel,
  Thirion, Grisel, Blondel, Prettenhofer, Weiss, Dubourg, Vanderplas, Passos,
  Cournapeau, Brucher, Perrot, and Duchesnay}]{scikit-learn}
Pedregosa, F.; Varoquaux, G.; Gramfort, A.; Michel, V.; Thirion, B.; Grisel,
  O.; Blondel, M.; Prettenhofer, P.; Weiss, R.; Dubourg, V.; Vanderplas, J.;
  Passos, A.; Cournapeau, D.; Brucher, M.; Perrot, M.; and Duchesnay, E. 2011.
\newblock Scikit-learn: Machine Learning in {P}ython.
\newblock \emph{Journal of Machine Learning Research}, 12: 2825--2830.

\bibitem[{Ribeiro, Singh, and Guestrin(2016)}]{ribeiro2016nothing}
Ribeiro, M.~T.; Singh, S.; and Guestrin, C. 2016.
\newblock Nothing else matters: model-agnostic explanations by identifying
  prediction invariance.
\newblock \emph{arXiv preprint arXiv:1611.05817}.

\bibitem[{Srivastava, Greff, and Schmidhuber(2015)}]{srivastava2015highway}
Srivastava, R.~K.; Greff, K.; and Schmidhuber, J. 2015.
\newblock Highway networks.
\newblock \emph{arXiv preprint arXiv:1505.00387}.

\bibitem[{Su et~al.(2015)Su, Wei, Varshney, and
  Malioutov}]{su2015interpretable}
Su, G.; Wei, D.; Varshney, K.~R.; and Malioutov, D.~M. 2015.
\newblock Interpretable two-level boolean rule learning for classification.
\newblock \emph{arXiv preprint arXiv:1511.07361}.

\bibitem[{Sundararajan, Taly, and Yan(2017)}]{sundararajan2017axiomatic}
Sundararajan, M.; Taly, A.; and Yan, Q. 2017.
\newblock Axiomatic attribution for deep networks.
\newblock In \emph{International Conference on Machine Learning}, 3319--3328.
  PMLR.

\bibitem[{Wisdom et~al.(2016)Wisdom, Powers, Pitton, and
  Atlas}]{wisdom2016interpretable}
Wisdom, S.; Powers, T.; Pitton, J.; and Atlas, L. 2016.
\newblock Interpretable recurrent neural networks using sequential sparse
  recovery.
\newblock \emph{arXiv preprint arXiv:1611.07252}.

\bibitem[{Zeiler and Fergus(2014)}]{zeiler2014visualizing}
Zeiler, M.~D.; and Fergus, R. 2014.
\newblock Visualizing and understanding convolutional networks.
\newblock In \emph{European conference on computer vision}, 818--833. Springer.

\bibitem[{Zeiler et~al.(2010)Zeiler, Krishnan, Taylor, and
  Fergus}]{zeiler2010deconvolutional}
Zeiler, M.~D.; Krishnan, D.; Taylor, G.~W.; and Fergus, R. 2010.
\newblock Deconvolutional networks.
\newblock In \emph{2010 IEEE Computer Society Conference on computer vision and
  pattern recognition}, 2528--2535. IEEE.

\bibitem[{Zeiler, Taylor, and Fergus(2011)}]{zeiler2011adaptive}
Zeiler, M.~D.; Taylor, G.~W.; and Fergus, R. 2011.
\newblock Adaptive deconvolutional networks for mid and high level feature
  learning.
\newblock In \emph{2011 International Conference on Computer Vision},
  2018--2025. IEEE.

\bibitem[{Zhou et~al.(2021)Zhou, Zhang, Peng, Zhang, Li, Xiong, and
  Zhang}]{zhou2021informer}
Zhou, H.; Zhang, S.; Peng, J.; Zhang, S.; Li, J.; Xiong, H.; and Zhang, W.
  2021.
\newblock Informer: Beyond efficient transformer for long sequence time-series
  forecasting.
\newblock In \emph{Proceedings of AAAI}.

\end{thebibliography}
\end{document}